\documentclass{article}

\usepackage{PRIMEarxiv}

\usepackage[utf8]{inputenc} 
\usepackage[T1]{fontenc}    
\usepackage{hyperref}       
\usepackage{url}            
\usepackage{booktabs}       
\usepackage{amsfonts}       
\usepackage{nicefrac}       
\usepackage{microtype}      
\usepackage{lipsum}
\usepackage{fancyhdr}       
\usepackage{graphicx}       
\graphicspath{{media/}}     
\usepackage{filecontents}
\usepackage{amsmath,amssymb,amsfonts}
\usepackage{algorithmic}
\usepackage{textcomp}
\usepackage{xcolor}
\usepackage[most]{tcolorbox}
\usepackage{framed}
\definecolor{shadecolor}{RGB}{0,200,230}

\title{Chatmap : Large Language Model Interaction with Cartographic Data

}

\author{
  Eren Unlu \\
  Datategy SAS \\
  Paris, France\\
  \texttt{eren.unlu@datategy.fr} \\
   \\
}

\begin{document}
\maketitle

\begin{abstract}
The swift advancement and widespread availability of foundational Large Language Models (LLMs), complemented by robust fine-tuning methodologies, have catalyzed their adaptation for innovative and industrious applications. Enabling LLMs to recognize and interpret geospatial data, while offering a linguistic access to vast cartographic datasets, is of significant importance. OpenStreetMap (OSM) is the most ambitious open-source global initiative offering detailed urban and rural geographic data, curated by a community of over 10 million contributors, which constitutes a great potential for LLM applications. In this study, we demonstrate the proof of concept and details of the process of fine-tuning a relatively small scale (1B parameters) LLM with an artificial dataset curated by a more capable teacher model, in order to provide a linguistic interface to the OSM data of an arbitrary urban region. Through this interface, users can inquire about a location's attributes, covering a wide spectrum of concepts, such as its touristic appeal or the potential profitability of various businesses in that vicinity. The study aims to provide an initial guideline for such generative artificial intelligence (AI) adaptations and demonstrate early signs of useful emerging abilities in this context even in minimal computational settings. The embeddings of artificially curated prompts including OSM data are also investigated in detail, which might be instrumental for potential geospatially aware urban Retrieval Augmented Generation (RAG) applications.
\end{abstract}


\section{Introduction}

In recent years, the explosive growth in the capabilities and utilities of Large Language Models (LLMs) has brought forth a paradigm shift in how we interact with and leverage data \cite{hadi2023survey}\cite{naveed2023comprehensive}. Traditionally, extracting value from vast datasets, especially those with specialized content like cartographic data, required a combination of domain expertise, time-consuming analysis, and specialized tools. With the advent of LLMs, there is an enticing opportunity to simplify this extraction process, making it more accessible and intuitive. The concept of linguistically querying geospatial datasets presents a confluence of the advances in natural language processing (NLP) and the ubiquity of current open digital cartographic data.

The implications of such advancements are vast and transformational. Corporate officials with no technical background can seek insights about whether a commercial venture they are considering is viable in an area that they simply selected by clicking on a map. Tourists could ask about the historical significance of regions they plan to visit and automatic touristic itinerary generation applications powered by LLMs with linguistic interfaces can be developed. Policy makers can profit from such a framework to optimally plan infrastructure investments such as new lines of subway in more human centric and geospatially aware interactions. Therefore, integration of LLMs with OSM like cartographic data is crucial.

In this paper, we present a basic framework aligned with this goal requiring very minimal computational budget and human labeled data in order to constitute a pioneering guideline and demonstrate that such productive applications in future can be developed with a reasonable amount of effort and time. The central idea of the paper is to fine-tune a foundational language model to comprehend the OSM features in an arbitrary area and align the content with the human intentions. The OSM database contains a vast amount of highly variant attributes from detailed mappings of power lines, road networks, buildings, designated areas to any type of amenities such as cafes, schools, public drinking water fountains \cite{mooney2017review}\cite{mooney2012annotation}. Consequently, for minimal proof of concept demonstration, we have strictly limited the OSM content we use in this study.

In our exemplary study, without loss of generality, circular areas with a 300-meter radius in the most densely tagged districts of Istanbul, Turkey, were selected. Specific quantitative aspects of selected OSM attributes within these circular areas are described verbally, which we refer to as 'preprompts'. Using a foundational LLM as a competent teacher, various prompt-response pairs were generated to form a training dataset for our fine-tuned model. Details on preprompt construction and guidance from the teacher model for effective artificial dataset curation are provided in the article.

An unaligned, open source foundational model of approximately just 1 billion parameters to fine-tune which is effectively pretrained on vast datasets is preferred to demonstrate the development of such abilities with significantly low amounts of resources. Using Low Rank Adaptation (LORA) and 8-bit quantization the fine-tuning process is performed on a single, moderate GPU \cite{hu2021lora}\cite{liu2023emergent}.

Thanks to the effective training dataset and relatively advanced linguistic capabilities of the pretrained base LLM compared to its moderate size, our fine-tuned model shows early signs of emergent abilities in this context. For locations which were not included in the fine-tuning dataset the model is queried to answer concepts mostly abstained in the training. In addition to LLM development, we have also investigated the embeddings of curated prompts including OSM attributes, which reflects the latent structure of urban landscape.

We believe that the minimal framework represented in this paper can encourage researchers to develop much advanced cartographic data aware generative AI applications. The prospects of such potential paradigms are also discussed.

\section{OSM Data and Preprompts}

OSM contains an immense amount of detailed data about specific geolocations. For our minimal setting, we have limited the quantitative data to a few key urban details. The concept presented in this paper is to define a point within an urban environment and then consider a circular area around it. For simplicity, we consistently chose radii of 300 meters in this study. However, varying radii or alternative methods of area selection could be considered. Urban attributes within this circular area are then retrieved from the OSM database. These attributes are subsequently articulated verbally. This verbal articulation serves as a “pre-prompt,” establishing the context for the questions posed to the LLM. For this study's purposes, we have limited our scope to the more densely tagged central districts of Istanbul, Turkey.

Chosen urban data to be included in the preprompts are : 

\begin{itemize}
  \item Number of different types of amenities : Amenities are primary categories that denote facilities and services vital for both visitors and residents. Examples of amenities include but are not limited to schools, ATMs, banks, pharmacies, libraries, and bus stops. An amenity essentially highlights features that offer convenience or serve essential functions in a particular geographical area. The number of each type of amenity in the area is verbally described.

  \item Intersecting administrative divisions : OSM database includes boundaries of all levels of administrative divisions of that particular country of interest. In order to incorporate urban geolocation awareness in the fine-tuning process, we have included the names of administrative bodies intersecting this area. For this case, only “districts” and “provinces” are considered. Note that, all geolocations in this study are in the province of “Istanbul”.

  \item Number and Surface Area of Buildings : In-detail tagged areas include polygon representations of buildings. As a measure of dense urbanization we have included the number of buildings residing in the circular area and the percentage of building surface area to the total circular area.
  
  \item Landuses : Landuses are polygon areas in OSM database defined according to their usage such as “military”, “park”, “construction”, “commercial”, “industrial” etc. As tagged residential areas are most of the time lacking in the region, they are excluded in this study. The percentages of each type of landuse surface to the total area are verbally expressed. For the sake of efficient contextualisation only landuse areas exceeding 2\% of the total surface area are included.
  
  \item Leisures : Leisures are polygon areas where they are defined based on their recreational purposes such as “water park”, “stadium”, “marina” etc.  For the sake of efficient contextualisation only leisure areas exceeding 2\% of the total surface area are included.

  \item Roads and Railways : OSM offers a very detailed description of all types of land roads with their types such as “motorway”, “residential street”, “secondary road”, “pedestrian road” etc and railways also including tram and subway lines. We calculate the total length in meters of each types of roads and railways in the circular area and verbally express it.
\end{itemize}

For instance, the preprompt of an arbitrary area is as follows : 

\begin{snugshade*}
\noindent\texttt{This is a circular area of radius of 300 meters that intersects province(s) of İstanbul and district(s) of Fatih. There are 3 atm(s), 2 bank(s), 1 bureau\_de\_change(s), 18 cafe(s), 2 clinic(s), 1 court\_house(s), 2 dentist(s), 1 driving\_school(s), 2 events\_venue(s), 11 fast\_food(s), 1 guest\_house(s), 3 hospital(s), 11 parking(s), 33 pharmacy(s), 9 place\_of\_worship(s), 1 post\_office(s), 43 restaurant(s), 5 school(s), 1 shower(s). There are 525 buildings which cover 31\% of the total area.  It contains 289 meters of platform rail, 100 meters of footway road, 80 meters of pedestrian road, 44 meters of primary\_link road, 2786 meters of residential road, 283 meters of service road, 20 meters of steps road, 1005 meters of tertiary road, 62 meters of tertiary\_link road, 249 meters of unclassified road.}
\end{snugshade*}

Note that, native texts of OSM values are used taking into consideration that the advanced teacher model is able to process them properly (also familiar with OSM nomenclature) to generate proper prompt-answer pairs.

\section{Artificial Dataset Curation with a Teacher Model}

Using advanced publicly available LLMs to curate supervised datasets to train or fine-tune smaller models is a well established concept. \cite{taori2023alpaca} is one of the most widely known examples of such approaches. By leveraging properly the detailed information and cognitive power of these models one can generate a quasi-infinite number of datapoints with sufficient quality. We have used OpenAI ChatGPT 3.5-turbo \cite{chalkidis2023chatgpt} to generate prompt-answer pairs for given preprompts. Some examples of prompts queried to ChatGPT 3.5-turbo are as follows : 

\begin{snugshade*}
\noindent\texttt{I will give these types of preprompts and you will generate prompt-answer pairs in python list of dictionaries format. These prompts should be questions that businessmen, citizens, tourists would demand based on the data in the preprompt. Generate 50 prompt-answer pairs with very diverse topics. Important : Do not generate prompts that data in preprompt is not sufficient to answer !}
\end{snugshade*}

\begin{snugshade*}
\noindent\texttt{preprompt = 'This is a circular area of radius of 300 meters that intersects province(s) of İstanbul and district(s) of Fatih. There are 10 atm(s), 2 bank(s), 1 bar(s), 11 bench(s), 1 bureau\_de\_change(s), 22 cafe(s), 1 car\_rental(s), 1 clinic(s), 7 fast\_food(s), 5 fountain(s), 1 ice\_cream(s), 4 library(s), 1 money\_transfer(s), 1 motel(s), 5 parking(s), 7 pharmacy(s), 14 place\_of\_worship(s), 2 police(s), 1 post\_office(s), 64 restaurant(s), 1 school(s), 1 theatre(s), 6 toilets(s), 1 university(s), 2 waste\_basket(s), 1 waste\_disposal(s). There are 200 buildings which cover 26\% of the total area. It contains 112 meters of platform rail, 2911 meters of footway road, 730 meters of pedestrian road, 2227 meters of residential road, 236 meters of service road, 270 meters of steps road, 330 meters of tertiary road.'}
\end{snugshade*}

\begin{snugshade*}
\noindent\texttt{I will give these types of preprompts and you will generate prompt-answer pairs in python list of dictionaries format. These prompts should be questions that businessmen, citizens, tourists would demand based on the data in the preprompt. Generate 50 prompt-answer pairs with very diverse topics. Important : Do not generate prompts that data in preprompt is not sufficient to answer !}
\end{snugshade*}

\begin{snugshade*}
\noindent\texttt{Generate 50 prompt-answer pairs, but be creative. Try to cover very different aspects in prompts, such as which type of commercial venture would suit here, whether it is residential or touristic, how you can describe this area etc. }
\end{snugshade*}

\begin{snugshade*}
\noindent\texttt{Important Warning : "Do not include questions that we dont have sufficient data in preprompt to answer" Before generating, repeat this last Important warning i gave, for affirmation. So, dont generate answers like "the preprmpt does not provide sufficient info ..." 
}
\end{snugshade*}

\begin{snugshade*}
\noindent\texttt{Topics shall be extremely variant : what type of commercial venture can be opened here, does it look like a transportation hub, what type of an urban area is this (residential, commercial, touristic, bussiness, industrial etc.) , does it look like a place where a grocery shop would earn much, is there a tram line, does it look like a central quarter in the city, etc. etc. should be extremely variant. Note, in python list of dictionaries
}
\end{snugshade*}

Below is an example of a datapoint created from the prompt-answer pairs generated by the teacher model : 

\begin{snugshade*}
\noindent\texttt{
'Area : This is a circular area of radius of 300 meters that intersects province(s) of İstanbul and district(s) of Fatih. There are 18 atm(s), 2 bank(s), 2 bench(s), 4 bicycle\_parking(s), 6 bureau\_de\_change(s), 31 cafe(s), 1 clinic(s), 3 doctors(s), 8 fast\_food(s), 1 fire\_station(s), 8 fountain(s), 1 gallery(s), 2 ice\_cream(s), 25 library(s), 1 motorcycle\_parking(s), 9 parking(s), 5 pharmacy(s), 10 place\_of\_worship(s), 1 police(s), 5 post\_office(s), 3 pub(s), 4 public\_bath(s), 2 public\_building(s), 135 restaurant(s), 3 school(s), 1 social\_centre(s), 2 social\_facility(s), 2 telephone(s), 1 theatre(s), 6 toilets(s), 1 university(s), 5 vending\_machine(s), 3 waste\_basket(s), 1 waste\_disposal(s). There are 293 buildings which cover 25\% of the total area.  The area is covered by 5\% park. It contains 242 meters of platform rail, 10 meters of tram rail, 58 meters of construction road, 2942 meters of footway road, 1786 meters of pedestrian road, 2060 meters of residential road, 126 meters of service road, 51 meters of steps road, 618 meters of tertiary road. Question : Tell me about the options for cultural enthusiasts with a gallery nearby. Answer : Cultural enthusiasts can explore the gallery in this area.'
}
\end{snugshade*}

At the end of the curation process we have generated 4111 datapoints from 81 different preprompts corresponding to different geolocations in Istanbul’s selected districts.

\section{1B Model and Fine-Tuning Process}

Our primary goal is to demonstrate the proof of concept of such a paradigm that cognitive capacity and linguistic human interface of specialized LLMs with even very constrained budgets can be incorporated into geospatial data. For this purpose, a 1 billion parameter is selected. The unaligned model, “Falcon 1B RW” is pretrained sufficiently on a vast refined dataset, therefore despite its relatively low number of parameters retains plausible amount of information and interpretive capacity \cite{penedo2023refinedweb}\cite{jiang2023empirical}. 

The model has a context length of 2048 tokens which is much above the number of tokens in any curated prompt. The Low Rank Adaptation (LORA) technique has indeed brought about a significant transformation in the process of fine-tuning in the field of deep learning, particularly within the domain of large language models (LLM). It has opened up numerous new possibilities for domain adaptation \cite{hu2021lora}. LORA's foundation rests on the insight highlighted in \cite{chai2023int2}, which emphasizes that the weight matrices of pre-trained language models possess a low-rank dimension.

With the weight matrices of an LLM : $W_{0} \in R^{d \times k}$, the update (fine-tuning) can be simply formulated as $W_{0} + \Delta W = W_{0} + BA$, where $B \in R^{d \times r}$ , $A \in R^{r \times k}$, and the rank $r << min(d, k)$. These additional tensors are called adapters. Therefore, optimizing the additional low-rank update $\Delta W$ can be done with exceptional efficiency by focusing solely on updating A and B, as these matrices contain significantly fewer parameters compared to the original weights. In addition as W is being kept frozen, catastrophic forgetting is mitigated up to a degree, whilst linguistic capacity of the pretrained weights are retained. It is well observed that adding adapters only on query-key-value tensors in self-attention heads yields plausible results for domain adaptation, hence for the sake of further decreasing computational budget LORA adapters are only applied on these components \cite{sun2023exploring}.

Quantization is another trick to decrease computational burden of LLMs significantly at the expense of reasonable performance loss \cite{kwon2022alphatuning}. 8-bit quantization is used for our specialized model. A LORA rank of 16 with an alpha value of 32 is used (0.1 dropouts). Aligned with our minimalistic proof of concept the model is fine-tuned only for approximately 2 epochs. 99\% of the dataset is used for training, whilst only the rest is spared for measuring validation loss. 

\begin{figure}[h]
\centering
\label{fig:fig_1}
\includegraphics[width=0.7\linewidth]{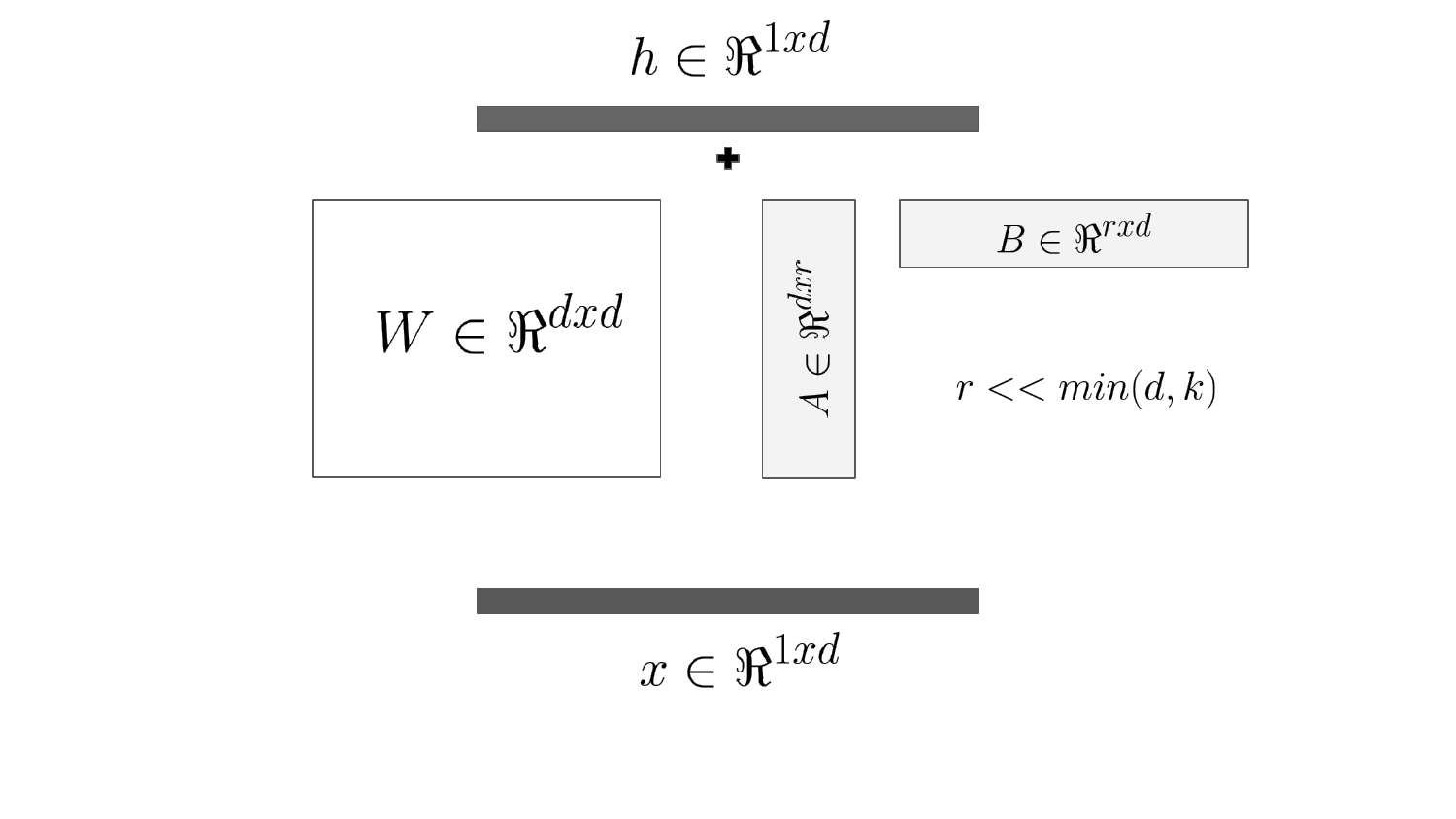}
\caption{LORA’s central idea is to inject additive low rank matrices to neural layers to update, whilst freezing the original weight matrix. This allows one to train a substantially lower number of parameters while avoiding catastrophic forgetting \cite{hu2021lora}.}
\end{figure}

\begin{figure}[h]
\centering
\label{fig:fig_2}
\includegraphics[width=0.65\linewidth]{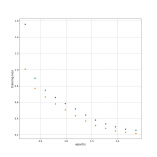}
\caption{Training (blue) and validation (orange) losses during approximately 2 epochs long fine-tuning process.}
\end{figure}

\section{Results}

The fine-tuned model is queried with different geolocations in the same districts with different types of questions in the training dataset and sometimes with different semantic context. Results suggest early signs of emerging abilities in this type of a paradigm and applicability of similar frameworks even with a very minimal amount of resources. Some examples of the responses received from the model are as follows :

\begin{snugshade*}
\noindent\texttt{
Area : This is a circular area of radius of 300 meters that intersects province(s) of İstanbul and district(s) of Ataşehir, Kadıköy. There are 10 atm(s), 1 bank(s), 4 bench(s), 2 cafe(s), 1 clinic(s), 1 dentist(s), 1 fast\_food(s), 1 fountain(s), 8 parking(s), 5 pharmacy(s), 1 place\_of\_worship(s), 3 restaurant(s), 3 school(s), 1 waste\_basket(s). There are 140 buildings which cover 18\% of the total area.  It contains 205 meters of footway road, 66 meters of pedestrian road, 944 meters of residential road, 532 meters of service road, 44 meters of tertiary road.
Question : How can we classify this region of Istanbul based on the data in prepromt? Explain in detail. 
Answer : Based on the data, this area appears to be a residential area with a mix of residential and commercial buildings. It could be classified as a residential neighborhood.
}
\end{snugshade*}

\begin{snugshade*}
\noindent\texttt{
Area : This is a circular area of radius of 300 meters that intersects province(s) of İstanbul and district(s) of Maltepe. There are 1 car\_wash(s), 1 drinking\_water(s), 4 pharmacy(s), 1 place\_of\_worship(s), 1 restaurant(s), 1 school(s), 2 taxi(s). There are 198 buildings which cover 22\% of the total area. The area is covered by 3\% grass.  It contains 1516 meters of residential road, 684 meters of tertiary road.
Question : Would it be convenient for a university student to live here? 
Answer : The presence of a university would be convenient for university students, but specific information about the student population is not available in the data.
}
\end{snugshade*}

\begin{snugshade*}
\noindent\texttt{
Area : This is a circular area of radius of 300 meters that intersects province(s) of İstanbul and district(s) of Beyoğlu. There are 2 atm(s), 3 bank(s), 1 cafe(s), 1 fast\_food(s), 1 fountain(s), 1 library(s), 2 parking(s), 8 pharmacy(s), 4 place\_of\_worship(s), 1 post\_office(s), 2 public\_bath(s), 1 recycling(s), 1 restaurant(s). There are 291 buildings which cover 28\% of the total area.  It contains 180 meters of primary road, 5417 meters of residential road, 24 meters of steps road, 37 meters of unclassified road.
Question : I am a father with a wife and two kids of age 4 and 6. Why would i consider to move here ?
Answer : With a family-friendly environment and a variety of amenities, this area could be a suitable place for a family to live.
}
\end{snugshade*}

\begin{snugshade*}
\noindent\texttt{
Area : This is a circular area of radius of 300 meters that intersects province(s) of İstanbul and district(s) of Kadıköy. There are 1 arts\_centre(s), 14 bank(s), 14 cafe(s), 2 dentist(s), 1 doctors(s), 5 fast\_food(s), 1 hospital(s), 4 ice\_cream(s), 1 library(s), 2 parking(s), 8 pharmacy(s), 1 post\_office(s), 1 pub(s), 11 restaurant(s), 1 veterinary(s). There are 157 buildings which cover 20\% of the total area.  It contains 68 meters of footway road, 104 meters of pedestrian road, 396 meters of primary road, 1013 meters of residential road, 14 meters of service road.
Question : Does it look like a place where a tourist would enjoy ? 
Answer : With cafes, restaurants, and a post office, it seems like a place where tourists would enjoy visiting.
}
\end{snugshade*}

\begin{snugshade*}
\noindent\texttt{
Area : This is a circular area of radius of 300 meters that intersects province(s) of İstanbul and district(s) of Kadıköy. There are 4 atm(s), 4 bank(s), 1 bar(s), 13 cafe(s), 2 clinic(s), 4 dentist(s), 1 doctors(s), 7 fast\_food(s), 3 ice\_cream(s), 1 kindergarten(s), 1 library(s), 2 parking(s), 1 parking\_entrance(s), 5 pharmacy(s), 1 place\_of\_worship(s), 1 post\_office(s), 2 pub(s), 16 restaurant(s), 1 veterinary(s). There are 140 buildings which cover 18\% of the total area.  It contains 128 meters of footway road, 174 meters of pedestrian road, 511 meters of primary road, 1154 meters of residential road, 502 meters of service road. 
Question : What type of people would enjoy living here ? 
Answer : With cafes, restaurants, and a library, this area seems to be a good place for people who enjoy socializing.
}
\end{snugshade*}

\begin{snugshade*}
\noindent\texttt{
Area : This is a circular area of radius of 300 meters that intersects province(s) of İstanbul and district(s) of Beyoğlu. There are 1 atm(s), 1 bank(s), 2 bench(s), 3 cafe(s), 1 community\_centre(s), 5 ferry\_terminal(s), 1 fountain(s), 1 fuel(s), 1 marketplace(s), 1 place\_of\_worship(s), 1 restaurant(s), 1 toilets(s). There are 26 buildings which cover 4\% of the total area. The area is covered by 7\% construction.  It contains 603 meters of platform rail, 60 meters of tram rail, 1351 meters of footway road, 287 meters of residential road, 154 meters of steps road. 
Question : Which age group of people would enjoy visiting here ? 
Answer : The presence of cafes and restaurants suggests that this area could be suitable for a variety of age groups.
}
\end{snugshade*}

\begin{snugshade*}
\noindent\texttt{
Area : This is a circular area of radius of 300 meters that intersects province(s) of İstanbul and district(s) of Şişli. There are 1 bank(s), 3 cafe(s), 4 fast\_food(s), 2 hospital(s), 13 pharmacy(s), 2 place\_of\_worship(s), 1 post\_office(s), 5 restaurant(s), 1 theatre(s). There are 106 buildings which cover 11\% of the total area.  It contains 83 meters of footway road, 72 meters of pedestrian road, 1233 meters of primary road, 157 meters of primary\_link road, 604 meters of residential road, 37 meters of secondary\_link road, 210 meters of service road, 78 meters of tertiary road. 
Question : I m a lawyer. Should i open an office in this area ? 
Answer : The presence of a bank and a hospital suggests a need for legal services. Opening an office may be a good option.
}
\end{snugshade*}

\section{Embeddings of Preprompts}

It is a convenient idea to inspect the embeddings of the curated preprompts to have an insight on the general attributes of various geolocations. Without loss of generality, we have used average GLOVE word embeddings \cite{pennington2014glove} and projected them on to two dimensional space using Universal Manifold Approximation and Projection (UMAP) algorithm \cite{mcinnes2018umap}. Fig. 3 shows the plot of projected UMAP values, where color channels are determined linearly proportional to these for illustrative purposes. Locations corresponding to these preprompt embeddings are displayed on the map with same color values in Fig. 4. As expected, even with basic embedding mechanisms and models, the latent structure of verbal descriptions of OSM data yields insightful patterns. Bright red colors indicate more touristic locations, dark red colors more indicate business/commercial districts and bright greenish colors indicate relatively empty spaces, residential areas. Note that, as mentioned previously this study is a preliminary attempt of a minimalistic proof of concept, thus much more complex frameworks can be imagined enabling creative Retrieval Augmented Generation (RAG) applications in this context.

\begin{figure}[h]
\centering
\label{fig:fig_umap_1}
\includegraphics[width=0.5\linewidth]{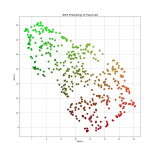}
\caption{Two dimensional UMAP values of various preprompts, where color values are linearly proportional to UMAP scores.}
\end{figure}

\begin{figure}[h]
\centering
\label{fig:fig_umap_2}
\includegraphics[width=0.65\linewidth]{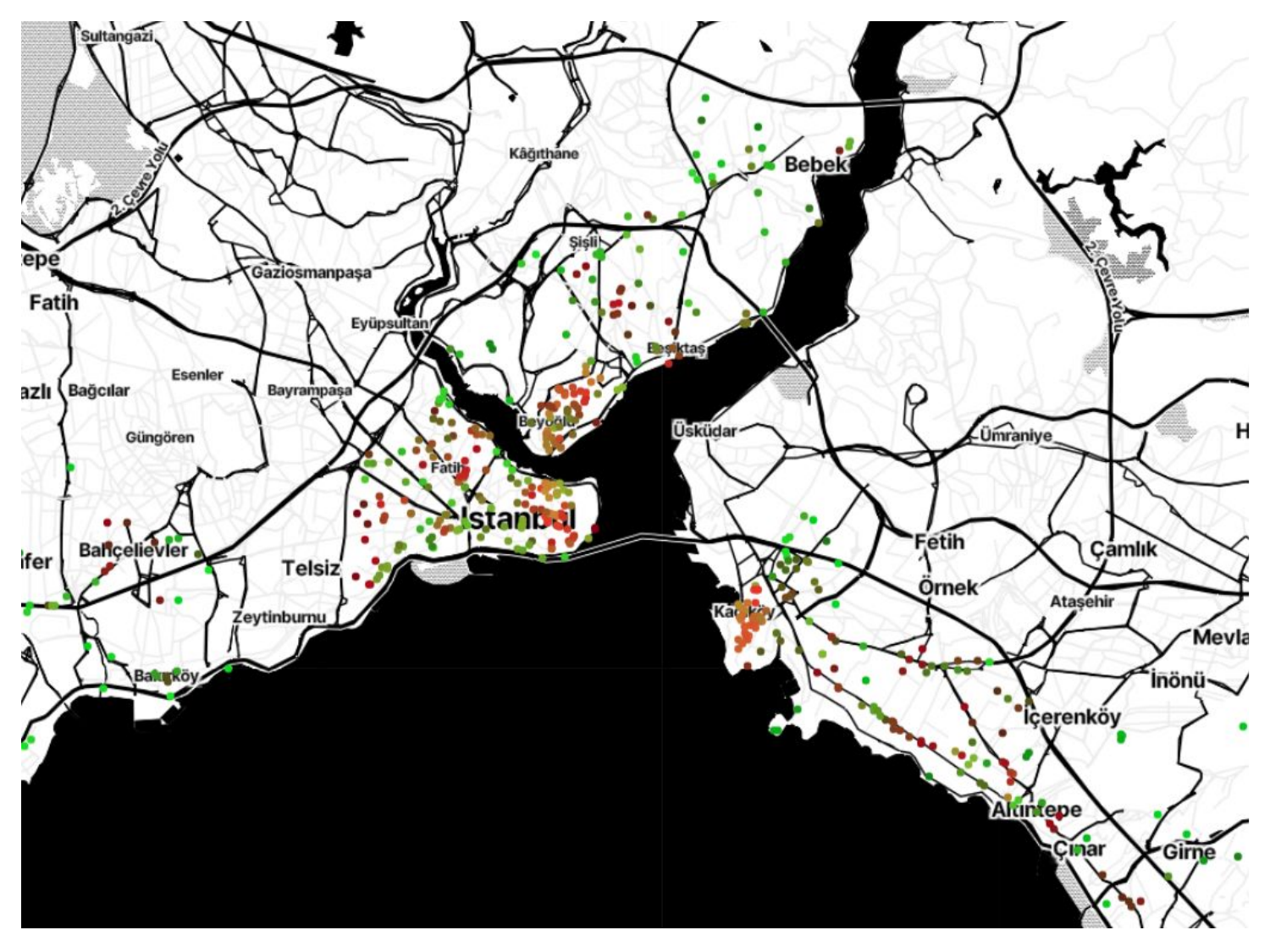}
\caption{Corresponding locations of the preprompts in Fig 3 with the same coloring.}
\end{figure}

\section{Conclusion and Prospects}

In this research, we explored the integration of LLMs with intricate cartographic datasets. Expressing OSM data of an urban region verbally and using these phrases as pre-context for queries, we have shown that such frameworks can be developed even with very minimal resources. Advanced publicly available LLMs can be used to generate artificial datasets automatically for this purpose such as in this study. Blending cartographic data with generative linguistic AI models offer vast amounts of possibilities.

\end{document}